\documentclass[11pt,a4paper]{article}
\usepackage[hyperref]{emnlp2020}
\usepackage{times}
\usepackage{latexsym}

\usepackage{microtype}
\usepackage{graphicx}
\usepackage{comment}
\usepackage{booktabs}
\usepackage{amssymb}
\usepackage{multirow}
\usepackage{multicol}

\definecolor{ourdarkgreen}{RGB}{84,130,53}
\definecolor{ourdarkblue}{RGB}{68,114,195}

\aclfinalcopy 
\def\aclpaperid{1159} 

\title{Learning from Context or Names?\\An Empirical Study on Neural Relation Extraction}

\author{Hao Peng$^{1}$\thanks{\quad Equal contribution}\hspace{0.5em}, Tianyu Gao$^{2*}$, Xu Han$^{1}$, Yankai Lin$^{3}$, Peng Li$^{3}$, Zhiyuan Liu$^{1}$\thanks{\quad Corresponding author e-mail: liuzy@tsinghua.edu.cn}\hspace{0.5em},  \\\textbf{Maosong Sun$^{1}$, Jie Zhou$^{3}$}\\
$^1$Department of Computer Science and Technology, Tsinghua University, Beijing, China\\
$^2$Princeton University, Princeton, NJ, USA\\
$^3$Pattern Recognition Center, WeChat AI, Tencent Inc, China\\
{\tt \{h-peng17,hanxu17\}@mails.tsinghua.edu.cn, tianyug@princeton.edu}
}

\date{}

\begin{document}
\maketitle
\begin{abstract}

Neural models have achieved remarkable success on relation extraction (RE) benchmarks. However, there is no clear understanding which type of information affects existing RE models to make decisions and how to further improve the performance of these models. To this end, we empirically study the effect of two main information sources in text: \textbf{textual context} and \textbf{entity mentions (names)}. We find that (i) while context is the main source to support the predictions, RE models also heavily rely on the information from entity mentions, most of which is type information, and (ii) existing datasets may leak shallow heuristics via entity mentions and thus contribute to the high performance on RE benchmarks. Based on the analyses, we propose an entity-masked contrastive pre-training framework for RE to gain a deeper understanding on both textual context and type information while avoiding rote memorization of entities or use of superficial cues in mentions. We carry out extensive experiments to support our views, and show that our framework can improve the effectiveness and robustness of neural models in different RE scenarios. All the code and datasets are released at \url{https://github.com/thunlp/RE-Context-or-Names}.

\end{abstract}

\section{Introduction}

\begin{figure}[t]
    \centering
    \includegraphics[width=0.94\linewidth]{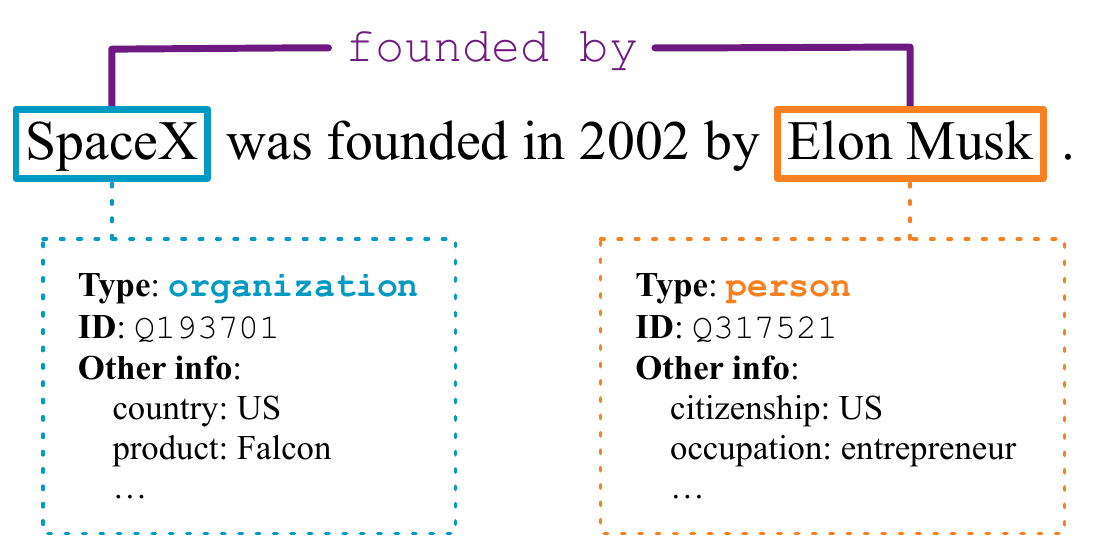}
    \caption{An example for the information provided by textual context and entity mentions in a typical RE scenario. 
    From mentions, we can acquire type information and link entities to KGs, and access further knowledge about them.
    The IDs in the figure are from Wikidata.}
    \label{fig:re_example}
\end{figure}

Relation extraction (RE) aims at extracting relational facts between entities from text, e.g., extracting the fact (SpaceX, \texttt{founded by}, Elon Musk) from the sentence in Figure~\ref{fig:re_example}. 
Utilizing the structured knowledge captured by RE, we can construct or complete knowledge graphs (KGs), and eventually support downstream applications like question answering~\citep{bordes2014question}, dialog systems~\citep{madotto-etal-2018-mem2seq} and search engines~\citep{xiong2017explicit}. With the recent advance of deep learning, neural relation extraction (NRE) models~\citep{socher2012semantic,liu2013convolution,soares2019matching} have achieved the latest state-of-the-art results and some of them are even comparable with human performance on several public RE benchmarks.

The success of NRE models on current RE benchmarks makes us wonder \textit{which type of information these models actually grasp to help them extract correct relations}. The analysis of this problem may indicate the nature of these models and reveal their remaining problems to be further explored. Generally, in a typical RE setting, there are two main sources of information in text that might help RE models classify relations: textual context and entity mentions (names). 


From human intuition, \textbf{textual context} should be the main source of information for RE. Researchers have reached a consensus that there exist interpretable patterns in textual context that express relational facts. For example, in Figure~\ref{fig:re_example}, ``\emph{... be founded ... by ...}'' is a pattern for the relation \texttt{founded by}. The early RE systems~\cite{huffman1995learning,califf-mooney-1997-relational} formalize patterns into string templates and determine relations by matching these templates. The later neural models~\cite{socher2012semantic,liu2013convolution} prefer to encode patterns into distributed representations and then predict relations via representation matching. Compared with rigid string templates, distributed representations used in neural models are more generalized and perform better.

Besides, \textbf{entity mentions} also provide much information for relation classification. As shown in Figure \ref{fig:re_example}, we can acquire the types of entities from their mentions, which could help to filter out those impossible relations. Besides, if these entities can be linked to KGs, models can introduce external knowledge from KGs to help RE~\citep{zhang2019ernie,peters-etal-2019-knowledge}. Moreover, for pre-trained language models, which are widely adopted for recent RE models, there may be knowledge about entities inherently stored in their parameters after pre-training~\citep{petroni-etal-2019-language}.


In this paper, we carry out extensive experiments to study to what extent RE models rely on the two information sources. We find out that:

(1) Both context and entity mentions are crucial for RE. As shown in our experiments, while context is the main source to support classification, entity mentions also provide critical information, most of which is the type information of entities. 

(2) Existing RE benchmarks may leak shallow cues via entity mentions, which contribute to the high performance of existing models. Our experiments show that models still can achieve high performance only given entity mentions as input, suggesting that there exist biased statistical cues from entity mentions in these datasets. 



The above observations demonstrate how existing models work on RE datasets, and suggest a way to further improve RE models: we should enhance them via better understanding context and utilizing entity types, while preventing them from simply memorizing entities or exploiting biased cues in mentions. 
From these points, we investigate an entity-masked contrastive pre-training framework for RE. 
We use Wikidata to gather sentences that may express the same relations, and let the model learn which sentences are close and which are not in relational semantics by a contrastive objective. 
In this process, we randomly mask entity mentions to avoid being biased by them. 
We show its effectiveness across several settings and benchmarks, and suggest that better pre-training technique is a reliable direction towards better RE. 


\section{Pilot Experiment and Analysis}
\label{sec:pilotexp}

To study which type of information affects existing neural RE models to make decisions, we first introduce some preliminaries of RE models and settings and then conduct pilot experiments as well as empirical analyses in this section. 

\subsection{Models and Dataset}
\label{sec:pilot_model_and_dataset}

There are various NRE models proposed in previous work (refer to Section~\ref{sec:relatedwork}), and we select the following three representative neural models for our pilot experiments and analyses:
\vspace{-0.5em}
\paragraph{CNN} We use the convolutional neural networks described in~\citet{nguyen2015relation} and augment the inputs with part-of-speech, named entity recognition and position embeddings following~\citet{zhang2017position}. 
\vspace{-0.5em}
\paragraph{BERT} BERT is a pre-trained language model that has been widely used in NLP tasks. We use BERT for RE following~\citet{soares2019matching}. In short, we highlight entity mentions in sentences by special markers and use the concatenations of entity representations for classification.

\vspace{-0.5em}
\paragraph{Matching the blanks (MTB)} MTB~\citep{soares2019matching} is an RE-oriented pre-trained model based on BERT. It is pre-trained by classifying whether two sentences mention the same entity pair with entity mentions randomly masked. It is fine-tuned for RE in the same way as BERT. Since it is not publicly released, we pre-train a BERT$_{\texttt{base}}$ version of MTB and give the details in Appendix~\ref{appendix:pretraining}.

There are also a number of public benchmarks for RE, and we select the largest supervised RE dataset \textbf{TACRED}~\citep{zhang2017position} in our pilot experiments. TACRED is a supervised RE dataset with $106,264$ instances and $42$ relations, which also provides type annotations for each entity. 

Note that we use more models and datasets in our main experiments, of which we give detailed descriptions and analyses in Section~\ref{sec:exp_and_ana}.


\subsection{Experimental Settings}

We use several input formats for RE, based on which we can observe the effects of context and entity mentions in controllable experiments. The following two formats are adopted by previous literature and are close to the real-world RE scenarios:

\vspace{-0.5em}
\paragraph{Context+Mention (C+M)} This is the most widely-used RE setting, where the whole sentence (with both context and highlighted entity mentions) is provided. To let the models know where the entity mentions are, we use position embeddings~\citep{zeng2014relation} for the CNN model and special entity markers~\citep{zhang2019ernie,soares2019matching} for the pre-trained BERT.

\vspace{-0.5em}
\paragraph{Context+Type (C+T)} We replace entity mentions with their types provided in TACRED. We use special tokens to represent them: for example, we use \texttt{[person]} and \texttt{[date]} to represent an entity with type \texttt{person} and \texttt{date} respectively. Different from \citet{zhang2017position}, we do not repeat the special tokens for entity-length times to avoid leaking entity length information.

\vspace{-0.5em}
\paragraph{} Besides the above settings, we also adopt three synthetic settings to study how much information context or mentions contribute to RE respectively:

\vspace{-0.5em}
\paragraph{Only Context (OnlyC)} To analyze the contribution of textual context to RE, we replace all entity mentions with the special tokens \texttt{[SUBJ]} and \texttt{[OBJ]}. In this case, the information source of entity mentions is totally blocked. 

\vspace{-0.5em}
\paragraph{Only Mention (OnlyM)} In this setting, we only provide entity mentions and discard all the other textual context for the input. 

\vspace{-0.5em}
\paragraph{Only Type (OnlyT)} This is similar to \textbf{OnlyM}, except we only provide entity types in this case. 

\begin{table}[t]
    \centering
    \small
    \begin{tabular}{l|ccccc}
    \toprule
    \textbf{Model} & \textbf{C+M}  & \textbf{C+T} & \textbf{OnlyC} & \textbf{OnlyM} & \textbf{OnlyT} \\
    \midrule
    CNN  & 0.547 & 0.591 & 0.441 & 0.434 & 0.295 \\
    BERT   & 0.683 & 0.686 & 0.570 & \textbf{0.466} & 0.277 \\
    MTB & \textbf{0.691} & \textbf{0.696} & \textbf{0.581} & 0.433 & \textbf{0.304}\\
    \bottomrule
    \end{tabular}
    \caption{TACRED results (micro F$_1$) with CNN, BERT and MTB on different settings.
    }
    \label{tab:supervised_tacred}
    \vspace{-1em}
\end{table}

\subsection{Result Analysis}

Table~\ref{tab:supervised_tacred} shows a detailed comparison across different input formats and models on TACRED. From the results we can see that:

(1) Both textual context and entity mentions provide critical information to support relation classification, and the most useful information in entity mentions is type information. As shown in Table~\ref{tab:supervised_tacred}, OnlyC, OnlyM and OnlyT suffer a significant performance drop compared to C+M and C+T, indicating that relying on only one source is not enough, and both context and entity mentions are necessary for correct prediction.  
Besides, we also observe that C+T achieves comparable results on TACRED with C+M for BERT and MTB. This demonstrates that most of the information provided by entity mentions is their type information. We also provide several case studies in Section~\ref{sec:exp_deeplook}, which further verify this conclusion.


\begin{table}[t]
    \centering
    \small
    \begin{tabular}{p{0.93\linewidth}}
    \toprule
          \multicolumn{1}{c}{\textbf{C+M}}\\
    \midrule
             Although her family was from Arkansas, \textit{\color{red}she} was born in \textit{\color{blue}Washington} state, where ... \\
         \textbf{\color{ourdarkgreen}Label}: \textbf{\texttt{per:state\_of\_birth}}\\
        \textbf{\color{ourdarkblue}Prediction}: \textbf{\texttt{per:state\_of\_residence}}\\
        
        \specialrule{0em}{4pt}{4pt}

           Dozens of lightly regulated subprime lenders, including New Century Financial Corp., have failed and troubled \textit{\color{red}Countrywide Financial Corp.} was acquired by \textit{\color{blue}Bank of America Corp.} \\
         \textbf{\color{ourdarkgreen}Label}: \textbf{\texttt{org:parents}}\\
        \textbf{\color{ourdarkblue}Prediction}: \textbf{\texttt{no\_relation}}\\
        
        \midrule
        \multicolumn{1}{c}{\textbf{C+T}}\\
        \midrule
          First, \textit{\color{blue}Natalie Hagemo} says, \textit{\color{red}she} fought the Church of Scientology just to give birth to her daughter. \\
        \textbf{\color{ourdarkgreen}Label}: \textbf{\texttt{no\_relation}}\\
        \textbf{\color{ourdarkblue}Prediction}: \textbf{\texttt{per:children}}\\
        
        \specialrule{0em}{4pt}{4pt}

         Earlier this week Jakarta hosted the \textit{\color{blue}general assembly} of the \textit{\color{red}Organisation of Asia-Pacific News Agencies}, ...\\
        \textbf{\color{ourdarkgreen}Label}: \textbf{\texttt{no\_relation}}\\
        \textbf{\color{ourdarkblue}Prediction}: \textbf{\texttt{org:members}}\\

        \specialrule{0em}{4pt}{4pt}

          The boy, identified by the Dutch foreign ministry as \textit{\color{blue}Ruben} but more fully by Dutch media as \textit{\color{red}Ruben van Assouw}, ...\\
         \textbf{\color{ourdarkgreen}Label}: \textbf{\texttt{per:alternate\_names}}\\
        \textbf{\color{ourdarkblue}Prediction}: \textbf{\texttt{no\_relation}}\\

    \bottomrule
    \end{tabular}
    \caption{Wrong predictions made only by C+M and only by C+T, where {\color{red}red} and {\color{blue}blue} represent subject and object entities respectively. As the examples suggest, C+M is more easily biased by the entity distribution in the training set and C+T loses some information from mentions that helps to understand the text.}
    \label{tab:cn_and_ct}
\end{table}

(2) There are superficial cues leaked by mentions in existing RE datasets, which may contribute to the high performance of RE models. We observe high performance on OnlyM with all three models on TACRED,
and this phenomenon also exists in other datasets (see Table~\ref{tab:supervised}).
We also take a deep look into the performance drop of OnlyC compared to C+M in Section~\ref{sec:exp_deeplook}, and find out that in some cases that models cannot well understand the context, they turn to rely on shallow heuristics from mentions. It inspires us to further improve models in extracting relations from context while preventing them from rote memorization of entity mentions.

\begin{table*}[h]
    \centering
    \small
    \begin{tabular}{lp{0.8\textwidth}}
    \toprule
        \textbf{Type} & \textbf{Example} \\
    \midrule
         Wrong  & ..., \textit{\color{blue}Jacinto Suarez}, Nicaraguan deputy to the \textit{\color{red}Central American Parliament} (PARLACEN) said Monday. \\
        $42\%$ &\textbf{\color{ourdarkgreen}Label}: \textbf{\texttt{org:top\_members/employees}}\\
        &\textbf{\color{ourdarkblue}Prediction}: \textbf{\texttt{no\_relation}}\\
        \specialrule{0em}{2pt}{2pt}
        & US life insurance giant MetLife said on Monday it will acquire \textit{\color{red}American International Group} unit American Life Insurance company (\textit{\color{blue}ALICO}) in a deal worth 155 billion dollars.\\
        &\textbf{\color{ourdarkgreen}Label}: \textbf{\texttt{org:subsidiaries}}\\
        &\textbf{\color{ourdarkblue}Prediction}: \textbf{\texttt{no\_relation}}\\
        \midrule
        No pattern & On Monday, the judge questioned the leader of the {\color{blue} \textit{Baptist}} group, {\color{red}\textit{Laura Silsby}}, who ... \\
         $31\%$ &\textbf{\color{ourdarkgreen}Label}: \textbf{\texttt{per:religion}}\\
        &\textbf{\color{ourdarkblue}Prediction}: \textbf{\texttt{no\_relation}}\\
        \midrule
        Confusing & About a year later, 
        \textit{\color{red}she} was transferred to Camp Hope, \textit{\color{blue}Iraq}.\\
         $27\%$ &\textbf{\color{ourdarkgreen}Label}: \textbf{\texttt{per:countries\_of\_residence}}\\
        &\textbf{\color{ourdarkblue}Prediction}: \textbf{\texttt{per:stateorprovinces\_of\_residence}}\\
    \bottomrule
    \end{tabular}
    \caption{Case study on unique wrong predictions made by OnlyC (compared to C+M). We sample $10\%$ of the wrong predictions, filter the wrong-labeled instances and manually annotate the wrong types to get the proportions. We use {\color{red}red} and {\color{blue}blue} to highlight the subject and object entities.}
    \label{tab:wrong}
\end{table*}

We notice that CNN results are a little inconsistent with BERT and MTB: CNN on OnlyC is almost the same as OnlyM, and C+M is $5\%$ lower than C+T. We believe that it is mainly due to the limited encoding power of CNN, which cannot fully utilize context and is more easily to overfit the shallow cues of entity mentions in the datasets.

\subsection{Case Study on TACRED}
\label{sec:exp_deeplook}

To further understand how performance varies on different input formats, we carry out a thorough case study on TACRED. We choose to demonstrate the BERT examples here because BERT represents the state-of-the-art class of models and we have observed a similar result on MTB.

First we compare C+M and C+T. We find out that C+M shares $95.7\%$ correct predictions with C+T, and $68.1\%$ wrong predictions of C+M are the same as C+T. It indicates that most information models take advantage of from entity mentions is their type information.
We also list some of the unique errors of C+M and C+T in Table~\ref{tab:cn_and_ct}. C+M may be biased by the entity distributions in the training set. For the two examples in Table~\ref{tab:cn_and_ct}, ``Washington'' is only involved in \texttt{per:stateorprovince\_of\_residence} and ``Bank of America Corp.'' is only involved in \texttt{no\_relation} in the training set, and this bias may  cause the error. On the other hand, C+T may have difficulty to correctly understand the text without specific entity mentions.
As shown in the example, after replacing mentions with their types, the model is confused by ``general assembly'' and fails to detect the relation between ``Ruben'' and ``Ruben van Assouw''. It suggests that entity mentions provide information other than types to help models understand the text. 

We also study why OnlyC suffers such a significant drop compared to C+M. In Table~\ref{tab:wrong}, we cluster all the unique wrong predictions made by OnlyC (compared to C+M) into three classes. ``Wrong'' represents sentences with clear patterns but misunderstood by the model. ``No pattern'' means that after masking the entity mentions, it is hard to tell what relation it is even for humans. ``Confusing'' indicates that after masking the entities, the sentence becomes ambiguous (e.g., confusing cities and countries). As shown in Table~\ref{tab:wrong}, in almost half (42\%) of the unique wrong predictions of OnlyC, the sentence has a clear relational pattern but the model fails to extract it, which suggests that in C+M, the model may rely on shallow heuristics from entity mentions to correctly predict the sentences. In the rest cases, entity mentions indeed provide critical information for classification.






\begin{figure*}[t]
    \centering
    \includegraphics[width=0.98\textwidth]{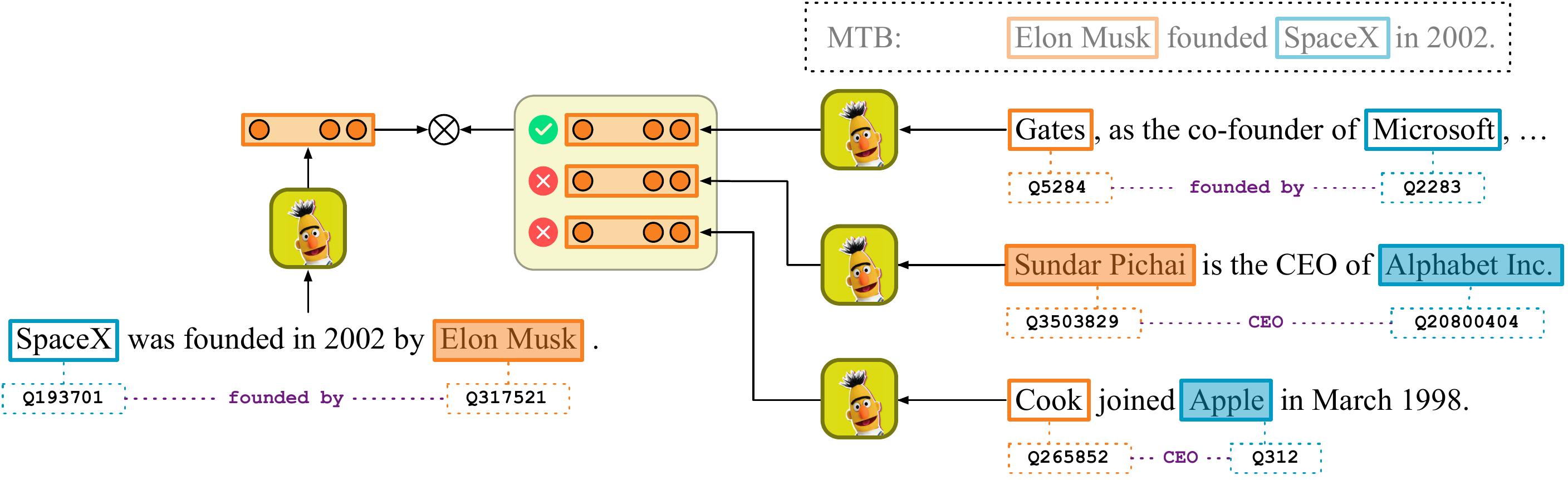}
    \caption{Our contrastive pre-training framework for RE. We assign relations to sentences by linking entity pairs in sentences to Wikidata and checking their relations in the KG. We assume that sentences with the same relation should have similar representations, and those with different relations should be pushed apart. Entity mentions are randomly masked (boxes with colored background) to avoid simple memorization.  Compared to MTB (in the dotted box), our method samples data with better diversity, which can not only increase the coverage of entity types and diverse context but also reduce the possibility of memorizing entity names.
}
    \label{fig:framework}
\end{figure*}

\section{Contrastive Pre-training for RE}
\label{sec:contrastive}

From the observations in Section \ref{sec:pilotexp}, we know that both context and entity type information is beneficial for RE models. However, in some cases RE models cannot well understand the relational patterns in context and rely on the shallow cues of entity mentions for classification. In order to enhance the ability to grasp entity types and extract relational facts from context, 
we propose the entity-masked contrastive pre-training framework for RE. We start with the motivation and process of relational contrastive example generation, and then go through the pre-training objective details.

\subsection{Relational Contrastive Example Generation}



We expect that by pre-training specifically towards RE, our model can be more effective at encoding relational representations from textual context and modeling entity types from mentions. 
To do so, we adopt the idea of contrastive learning~\cite{hadsell2006dimensionality}, which aims to learn representations by pulling ``neighbors'' together and pushing ``non-neighbors'' apart. After this, ``neighbor'' instances will have similar representations. 
So it is important to define ``neighbors'' in contrastive learning and we utilize the information from KGs to to that. Inspired by distant supervision~\citep{mintz2009distant}, we assume that sentences with entity pairs sharing the same relation in KGs are ``neighbors''.




Formally, denote the KG we use as $\mathcal{K}$, which is composed of relational facts. Denote two random sentences as $X_A$ and $X_B$, which have entity mentions $h_A, t_A$ and $h_B, t_B$ respectively. 
We define $X_A$ and $X_B$ as ``neighbors'' if there is a relation $r$ such that $(h_A, r, t_A)\in \mathcal{K}$ and $(h_B, r, t_B)\in \mathcal{K}$. 
We take Wikidata as the KG since it can be easily linked to the Wikipedia corpus used for pre-training. When training, we first sample a relation $r$ with respect to its proportion in the KG, and then sample a sentence pair ($X_A$, $X_B$) linked to $r$. To learn contrastively, we randomly sample $N$ sentences $X_B^i,1\leq i\leq N$ so they can form $N$ negative pairs with $X_A$. The model needs to classify which sentence among all the postive and negative samples has the same relation with $X_A$.


To avoid memorizing entity mentions or extracting shallow features from them during pre-training, we randomly mask entity mentions with the special token \texttt{[BLANK]}. We use $P_{\texttt{BLANK}}$ to denote the ratio of replaced entities and set $P_{\texttt{BLANK}}=0.7$ following~\citet{soares2019matching}. Note that masking all mentions during pre-training is also not a good option since it will create a gap between pre-training and fine-tuning and also block the pre-trained models from utilizing entity mention information (e.g., learning entity types). 

Take an example to understand our data generation process: In Figure~\ref{fig:framework}, there are two sentences ``\emph{\underline{SpaceX} was founded in 2002 by \underline{Elon Musk}}'' and ``\emph{As the co-founder of \underline{Microsoft}, \underline{Bill Gates} ...}'' where both (SpaceX, \texttt{founded by}, Elon Musk) and (Microsoft, \texttt{founded by}, Bill Gates) exist in the KG. We expect the two sentences to have similar representations reflecting the relation. On the other hand, for the other two sentences in the right part of the figure, since their entity pairs do not have the relation \texttt{founded by}, they are regarded as negative samples and are expected to have diverse representations from the left one. During pre-training, each entity mention has a probability of $P_{\texttt{BLANK}}$ to be masked.

The main problem of the generation process is that the sentence may express no relation between the entities at all, or express the relation different from what we expect. 
For example, a sentence mentioning ``SpaceX'' and ``Elon Musk'' may express the relation \texttt{founded by}, \texttt{CEO} or \texttt{CTO}, or simply does not express any relation between them. An example could be ``\textit{Elon Musk answers reporters' questions on a SpaceX press conference}'', which expresses no clear relation between the two. 
However, we argue that the noise problem is not critical for our pre-training framework: Our goal is to get relatively better representations towards RE compared to raw pre-trained models like BERT, rather than to directly train an RE model for downstream tasks, so noise in the data is acceptable.



With the help of the generated relational contrastive examples, our model can learn to better grasp type information from mentions and extract relational semantics from textual context: (1) The paired two sentences, which mention different entity pairs but share the same relation, prompt the model to discover the connections between these entity mentions for the relation. Besides, the entity masking strategy can effectively avoid simply memorizing entities. This eventually encourages the model to exploit entity type information. (2) Our generation strategy provides a diverse set of textual context expressing the same relation to the model, which motivates the model to learn to extract the relational patterns from a variety of expressions. 

Compared with our model, MTB~\citep{soares2019matching} takes a more strict rule which requires the two sampled sentences to share the same entity pair. While it reduces the noise, the model also samples data with less diversity and loses the chance to learn type information. 

\subsection{Training Objectives}
\label{sec:contrastive_obj}

\begin{table}[t]
    \centering
    \small
    \begin{tabular}{lrrr}
        \toprule
        \textbf{Dataset} & \textbf{\#~Rel.} & \textbf{\#~Inst.} & \textbf{\%~N/A}\\
        \midrule
        TACRED & 42 & 106,264 & 79.5\% \\
        SemEval-2010 Task 8 & 19 & 10,717 & 17.4\% \\
        Wiki80 & 80 & 56,000 & - \\
        ChemProt & 13 & 10,065 & - \\
        FewRel & 100 & 70,000 & -\\
        \bottomrule
    \end{tabular}
    \caption{Statistics for RE datasets used in the paper, including numbers of relations, numbers of instances and proportions of \texttt{N/A} instances. ``-'' for the last column means that there is no \texttt{N/A} relation in the dataset.}
    \label{tab:sup_dataset}
\end{table}

In our contrastive pre-training, we use the same Transformer architecture~\citep{vaswani2017attention} as BERT. 
Denote the Transformer encoder as $\texttt{ENC}$ and the output at the position $i$ as $\texttt{ENC}_i(\cdot)$. For the input format, we use special markers to highlight the entity mentions following~\citet{soares2019matching}. For example, for the sentence ``\emph{\underline{SpaceX} was founded by \underline{Elon Musk}}.'', the input sequence is ``\texttt{[CLS]}\texttt{[E1]} SpaceX \texttt{[/E1]} was founded by \texttt{[E2]} Elon Musk \texttt{[/E2]} . \texttt{[SEP]}''.

During the pre-training, we have two objectives: contrastive pre-training objective and masked language modeling objective.

\vspace{-0.5em}
\paragraph{Contrastive Pre-training Objective} As shown in Figure~\ref{fig:framework}, given the positive sentence pair $(x_A, x_B)$, and negative sentence pairs $(x_A, x_B^i), 1\leq i\leq N$, we first use the Transformer encoder to get relation-aware representation for $x$ in $\{x_A,x_B\}\cup\{x_B^i\}_{i=1}^{N}$:

\begin{equation}
    \mathbf{x} = \texttt{ENC}_{h}(x)\oplus\texttt{ENC}_{t}(x),
\end{equation}
where $h$ and $t$ are the positions of special tokens \texttt{[E1]} and \texttt{[E2]}, and $\oplus$ stands for concatenation. 
With the sentence representation, we have the following training objective:

\begin{equation}
    \mathcal{L}_{CP}=-\log \frac{e^{\mathbf{x}_A^T\mathbf{x}_B}}{e^{\mathbf{x}_A^T\mathbf{x}_B}+\sum_{i=1}^{i\leq N}e^{\mathbf{x}_A^T\mathbf{x}_B^i}}.
\end{equation}

By optimizing the model with respect to $\mathcal{L}_{CP}$, we expect representations for $x_A$ and $x_B$ to be closer and eventually sentences with similar relations will have similar representations.

\vspace{-0.5em}
\paragraph{Masked Language Modeling Objective} To maintain the ability of language understanding inherited from BERT and avoid catastrophic forgetting \citep{mccloskey1989catastrophic}, we also adopt the masked language modeling (MLM) objective from BERT. MLM randomly masks tokens in the inputs and by letting the model predict the masked tokens, MLM learns contextual representation that contains rich semantic and syntactic knowledge. Denote the MLM loss as $\mathcal{L}_{MLM}$.


Eventually, we have the following training loss:
\begin{equation}
    \mathcal{L}=\mathcal{L}_{CP}+\mathcal{L}_{MLM}.
\end{equation}

\section{Experiment}
\label{sec:exp_and_ana}

In this section, we explore the effectiveness of our relational contrastive pre-training across two typical RE tasks and several RE datasets.




\subsection{RE Tasks}

\begin{table*}[h]
    \small
    \centering
    \begin{tabular}{l|l|ccc|ccc|ccc}
        \toprule
        \multirow{2}{*}{\textbf{Dataset}} & \multirow{2}{*}{\textbf{Model}} & \multicolumn{3}{c|}{\textbf{1\%}} & \multicolumn{3}{c|}{\textbf{10\%}} & \multicolumn{3}{c}{\textbf{100\%}}\\
        & & \textbf{C+M} & \textbf{OnlyC} & \textbf{OnlyM} & \textbf{C+M} & \textbf{OnlyC} & \textbf{OnlyM} & \textbf{C+M} & \textbf{OnlyC} & \textbf{OnlyM} \\
        \midrule
        \multirow{3}{*}{TACRED} & BERT & 0.211  & 0.167 & 0.220 & 0.579 & 0.446 & 0.433 & 0.683 & 0.570 & \textbf{0.466}\\
        & MTB & 0.304 & 0.231 & 0.308 & 0.608 & 0.496 & 0.441 & 0.691 & 0.581 & 0.433 \\
        & CP & \textbf{0.485} & \textbf{0.393} & \textbf{0.350} &  \textbf{0.633} & \textbf{0.515} & \textbf{0.453} & \textbf{0.695} & \textbf{0.593} & 0.450 \\
        \midrule
        \multirow{3}{*}{SemEval} & BERT & 0.367 & 0.294 & 0.245  & 0.772 & 0.688 & 0.527 & 0.871 & 0.798 & 0.677 \\
        & MTB & 0.362 & 0.330 & \textbf{0.249} & 0.806 & 0.744 & 0.543 &  0.873 & 0.807 & \textbf{0.682}\\
        & CP & \textbf{0.482}  & \textbf{0.470} & 0.221 & \textbf{0.822}  & \textbf{0.766}  & \textbf{0.543} &  \textbf{0.876}  & \textbf{0.811}  & 0.679 \\
        \midrule
        \multirow{3}{*}{Wiki80} & BERT & 0.559 & 0.413 & 0.463  & 0.829 & 0.413 & 0.655& 0.913 & 0.810 & 0.781\\
        & MTB & 0.585  & 0.509 & 0.542  & 0.859 & 0.509 & 0.719   & 0.916 & 0.820 & 0.788 \\
        & CP & \textbf{0.827}  & \textbf{0.734} & \textbf{0.653}  & \textbf{0.893}  & \textbf{0.734} & \textbf{0.745}  &  \textbf{0.922}  & \textbf{0.834} & \textbf{0.799}  \\
        \midrule
        \multirow{3}{*}{ChemProt} & BERT & 0.362  & 0.362 & 0.362  & 0.634  & 0.584 & 0.385 &  0.792  & 0.777 & 0.463\\
        & MTB & \textbf{0.362}  & 0.362 & \textbf{0.362}  & 0.682 & 0.685 & 0.403&  0.796  & 0.798 & 0.463  \\
        & CP & 0.361 & \textbf{0.362} & 0.360  & \textbf{0.708}  & \textbf{0.697} & \textbf{0.404} &  \textbf{0.806}  & \textbf{0.803} & \textbf{0.467}  \\
        \bottomrule
    \end{tabular}
    \caption{Results on supervised RE datasets TACRED~(micro F$_1$), SemEval~(micro F$_1$), Wiki80~(accuracy) and ChemProt~(micro F$_1$). 1\%~/~10\% indicate using 1\%~/~10\% supervised training data respectively. }
    \label{tab:supervised}
\end{table*}

For comprehensive experiments, we evaluate our models on various RE tasks and datasets.

\vspace{-0.5em}
\paragraph{Supervised RE} This is the most widely-adopted setting in RE, where there is a pre-defined relation set $\mathcal{R}$ and each sentence $x$ in the dataset expresses one of the relations in $\mathcal{R}$. In some benchmarks, there is a special relation named \texttt{N/A} or \texttt{no\_relation}, indicating that the sentence does not express any relation between the given entities, or their relation is not included in $\mathcal{R}$.

For supervised RE datasets, we use TACRED \citep{zhang2017position}, SemEval-2010 Task 8 \citep{hendrickx-etal-2009-semeval}, Wiki80 \citep{han-etal-2019-opennre} and ChemProt \citep{kringelum2016chemprot}. Table \ref{tab:sup_dataset} shows the comparison between the datasets. 

We also add $1\%$ and $10\%$ settings, meaning using only $1\%$ / $10\%$ data of the training sets. It is to simulate a low-resource scenario and observe how model performance changes across different datasets and settings. Note that ChemProt only has $4,169$ training instances, which leads to the abnormal results on $1\%$ ChemProt in Table~\ref{tab:supervised}. We give details about this problem in Appendix~\ref{appendix:fine-tuning}. 

\vspace{-0.5em}
\paragraph{Few-Shot RE} Few-shot learning is a recently emerged topic to study how to train a model with only a handful of examples for new tasks. A typical setting for few-shot RE is $N$-way $K$-shot RE \citep{han2018fewrel}, where for each evaluation episode, $N$ relation types, $K$ examples for each type and several query examples (all belonging to one of the $N$ relations) are sampled, and models are required to classify the queries based on given $N\times K$ samples. We take FewRel~\citep{han2018fewrel,gao-etal-2019-fewrel} as the dataset and list its statistics in Table~\ref{tab:sup_dataset}.

We use Prototypical Networks as in~\citet{snell2017prototypical,han2018fewrel} and make a little change: (1) We take the representation as described in Section~\ref{sec:contrastive_obj} instead of using \texttt{[CLS]}. (2) We use dot production instead of Euclidean distance to measure the similarities between instances. We find out that this method outperforms original Prototypical Networks in~\citet{han2018fewrel} by a large margin.


\subsection{RE Models}


\begin{table*}[t]
    \centering
    \small
    \scalebox{0.96}{
    \begin{tabular}{l|ccc|ccc|ccc|ccc}
    \toprule
        \multirow{2}{*}{\textbf{Model}} &  \multicolumn{3}{c|}{\textbf{5-way 1-shot}} & 
        \multicolumn{3}{c|}{\textbf{5-way 5-shot}} &
        \multicolumn{3}{c|}{\textbf{10-way 1-shot}} &
        \multicolumn{3}{c}{\textbf{10-way 5-shot}}\\
        & \textbf{C+M} & \textbf{OnlyC} & \textbf{OnlyM} & \textbf{C+M} & \textbf{OnlyC} & \textbf{OnlyM} & \textbf{C+M} & \textbf{OnlyC} & \textbf{OnlyM} & \textbf{C+M} & \textbf{OnlyC} & \textbf{OnlyM}  \\
        \midrule
        \multicolumn{13}{c}{FewRel 1.0}\\
        \midrule
        BERT&0.911 &0.866& 0.701 &0.946 &0.925 & 0.804 &0.842 & 0.779 &0.575 & 0.908 & 0.876 & 0.715\\
        MTB & 0.911&0.879&0.727 & 0.954& 0.939&0.835 &0.843 & 0.779&0.568&0.918 &0.892&0.742  \\
        CP  &\textbf{0.951} & \textbf{0.926} & \textbf{0.743} & \textbf{0.971} & \textbf{0.956} &\textbf{0.840} & \textbf{0.912} & \textbf{0.867} &\textbf{0.620}&\textbf{0.947} & \textbf{0.924}&\textbf{0.763} \\
        \midrule
        \multicolumn{13}{c}{FewRel 2.0 Domain Adaptation}\\
        \midrule
        BERT & 0.746&0.683 &0.316&0.827 &0.782 &0.406 &0.635 &0.542 &0.210&0.765 & 0.706&0.292\\
        MTB & 0.747& 0.692&\textbf{0.338}&\textbf{0.879} &0.836 &0.426 &0.625 &0.528 &\textbf{0.216}&\textbf{0.811} &\textbf{0.744} &\textbf{0.298} \\
        CP &\textbf{0.797} &\textbf{0.745} &0.335&0.849 &\textbf{0.840} &\textbf{0.437} &\textbf{0.681} &\textbf{0.601} &0.213&0.798 &0.738 &0.297\\
    \bottomrule
    \end{tabular}
    }
    \caption{Accuracy on FewRel dataset. FewRel 1.0 is trained and tested on Wikipedia domain. FewRel 2.0 is trained on Wikipedia domain but tested on biomedical domain.}
    \label{tab:fewshot}
\end{table*}


Besides BERT and MTB we have introduced in Section~\ref{sec:pilot_model_and_dataset}, we also evaluate our proposed \textbf{c}ontrastive \textbf{p}re-training framework for RE (CP). We write the detailed hyper-parameter settings of both the pre-training and fine-tuning process for all the models in Appendix~\ref{appendix:pretraining} and~\ref{appendix:fine-tuning}.

Note that since MTB and CP use Wikidata for pre-training, and Wiki80 and FewRel are constructed based on Wikidata, we exclude all entity pairs in test sets of Wiki80 and FewRel from pre-training data to avoid test set leakage.

\subsection{Strength of Contrastive Pre-training}

Table~\ref{tab:supervised}~and~\ref{tab:fewshot} show a detailed comparison between BERT, MTB and our proposed contrastive pre-trained models. Both MTB and CP improve model performance across various settings and datasets, demonstrating the power of RE-oriented pre-training. Compared to MTB, CP has achieved even higher results, proving the effectiveness of our proposed contrastive pre-training framework. To be more specific, we observe that:

(1) CP improves model performance on all C+M, OnlyC and OnlyM settings, indicating that our pre-training framework enhances models on both context understanding and type information extraction. 

(2) The performance gain on C+M and OnlyC is universal, even for ChemProt and FewRel~2.0, which are from biomedical domain. Our models trained on Wikipedia perform well on biomedical datasets, suggesting that CP learns relational patterns that are effective across different domains.

(3) CP also shows a prominent improvement of OnlyM on TACRED, Wiki80 and FewRel~1.0, which are closely related to Wikipedia. It indicates that our model has a better ability to extract type information from mentions. Both promotions on context and mentions eventually lead to better RE results of CP (better C+M results).

(4) The performance gain made by our contrastive pre-training model is more significant on low-resource and few-shot settings. For C+M, we observe a promotion of $7\%$ on 10-way 1-shot FewRel 1.0, $18\%$ improvement on $1\%$ setting of TACRED, and $24\%$ improvement on $1\%$ setting of Wiki80. There is also a similar trend for OnlyC and OnlyM. In the low resource and few-shot settings, it is harder for models to learn to extract relational patterns from context and easier to overfit to superficial cues of mentions, due to the limited training data. However, with the contrastive pre-training, our model can relatively take better use of textual context while avoiding being biased by entities, and outperform the other baselines by a large margin.




\section{Related Work}
\label{sec:relatedwork}

\paragraph{Development of RE} RE of early days has gone through pattern-based methods~\citep{huffman1995learning,califf-mooney-1997-relational}, feature-based methods~\citep{kambhatla2004combining,guodong2005exploring}, kernel-based methods~\citep{culotta2004dependency,bunescu2005shortest}, graphical models~\citep{roth2002probabilistic,roth2004linear}, etc. Since \citet{socher2012semantic} propose to use recursive neural networks for RE, there have been extensive studies on neural RE~\citep{liu2013convolution,zeng2014relation,zhang2015relation}. To solve the data deficiency problem, researchers have developed two paths: \textbf{distant supervision}~\citep{mintz2009distant,min2013distant,riedel2010modeling,zeng2015distant,lin2016neural} to automatically collect data by aligning KGs and text, and \textbf{few-shot learning}~\citep{han2018fewrel,gao-etal-2019-fewrel} to learn to extract new relations by only a handful of samples. 

\paragraph{Pre-training for RE} With the recent advance of pre-trained language models~\citep{devlin2019bert}, applying BERT-like models as the backbone of RE systems~\citep{soares2019matching} has become a standard procedure. Based on BERT, \citet{soares2019matching} propose matching the blanks, an RE-oriented pre-trained model to learn relational patterns from text. A different direction 
is to inject entity knowledge, in the form of entity embeddings, into BERT~\citep{zhang2019ernie,peters-etal-2019-knowledge,liu2019k}. We do not discuss this line of work here for their promotion comes from relational knowledge of external sources, while we focus on text itself in the paper. 

\paragraph{Analysis of RE} \citet{han2020data} suggest to study how RE models learn from context and mentions. \citet{alt2020tacred} also point out that there may exist shallow cues in entity mentions. However, there have not been systematical analyses about the topic  
and to the best of our knowledge, we are the first one to thoroughly carry out these studies.

\section{Conclusion}

In this paper, we thoroughly study how textual context and entity mentions affect RE models respectively. Experiments and case studies prove that (i) both context and entity mentions (mainly as type information) provide critical information for relation extraction, and (ii) existing RE datasets may leak superficial cues through entity mentions and models may not have the strong abilities to understand context as we expect. From these points, we propose an entity-masked contrastive pre-training framework for RE to better understand textual context and entity types, and experimental results prove the effectiveness of our method. 

In the future, we will continue to explore better RE pre-training techniques, especially with a focus on open relation extraction and relation discovery. These problems require models to encode good relational representation with limited or even zero annotations, and we believe that our pre-trained RE models will make a good impact in the area.


\section*{Acknowledgments}

This work is supported by the National Key Research and Development Program of China (No. 2018YFB1004503), the National Natural Science Foundation of China (NSFC No. 61532010) and Beijing Academy of Artificial Intelligence (BAAI). This work is also supported by the Pattern Recognition Center, WeChat AI, Tencent Inc. Gao is supported by 2019 Tencent Rhino-Bird Elite Training Program. Gao and Peng are both supported by Tsinghua University Initiative Scientific Research Program. 

\bibliography{emnlp2020}
\bibliographystyle{acl_natbib}

\newpage

\appendix

\section{Pre-training Details} 
\label{appendix:pretraining}

\paragraph{Pre-training Dataset} We construct a dataset for pre-training following the method in the paper. We use Wikipedia articles as corpus and Wikidata~\citep{vrandevcic2014wikidata} as the knowledge graph. Firstly, We 
use anchors to link entity mentions in Wikipedia corpus with entities in Wikidata. Then, in order to link more unanchored entity mentions, we adopt \texttt{spaCy}\footnote{\url{https://spacy.io/}} to find all possible entity mentions, and link them to entities in Wikidata via name matching. Finally, we get a pre-training dataset containing $744$ relations and $867,278$ sentences. We release this dataset together with our source code at our GitHub repository\footnote{\url{https://github.com/thunlp/RE-Context-or-Names}}.

We also use this dataset for MTB, which is slightly different from the original paper~\citep{soares2019matching}. The original MTB takes all entity pairs into consideration, even if they do not have a relationship in Wikidata. Using the above dataset means that we filter out these entity pairs. We do this out of training efficiency, for those entity pairs that do not have a relation are likely to express little relational information, and thus contribute little to the pre-training.

\paragraph{Data Sampling Strategy} For MTB~\citep{soares2019matching}, we follow the same sampling strategy as in the original paper.
For pre-training our contrastive model, we regard sentences labeled with the same relation as a ``bag''. Any sentence pair whose sentences are in the same bag is treated as a positive pair and as a negative pair otherwise. So there will be a large amount of possible positive samples and negative samples. We dynamically sample positive pairs of a relation with respect to the number of sentences in the bag. 

\paragraph{Hyperparameters} We use Huggingface's \texttt{Transformers}\footnote{\url{https://github.com/huggingface/transformers}} to implement models for both pre-training and fine-tuning and use AdamW~\citep{loshchilov2018fixing} for optimization. 
For most pre-training hyperparameters, we select the same values as \citet{soares2019matching}. We search hyperparameter batch size in \{$256$, $2048$\} and $P_\texttt{BLANK}$ in \{$0.3$, $0.7$\}. For MTB, batch size $N$ means that a batch contains $2N$ sentences, which form $N/2$ positive pairs and $N/2$ negative pairs. For CP, batch size $N$ means that a batch contains $2N$ sentences, which form $N$ positive pairs.
For negative samples, we pair the sentence in each pair with sentences in other pairs. 

We set hyperparameters according to results on supervised RE dataset TACRED (micro F$_1$).
Table \ref{tab:params} shows hyperparameters for pre-training MTB and our contrastive model (CP). The batch size of our implemented MTB is different from that in \citet{soares2019matching}, because in our experiments, MTB with a batch size of $256$ performs better on TACRED than the batch size of $2048$.

\begin{table}[]
\small
    \centering
    \begin{tabular}{l|cc}
        \toprule
        \textbf{Parameter} &  \textbf{MTB} & \textbf{CP} \\
        \midrule 
        Learning Rate & $3\times10^{-5}$ & $3\times10^{-5}$\\
        Batch Size & $256$ & $2048$ \\
        Sentence Length & $64$ & $64$ \\
        $P_{\texttt{BLANK}}$ & $0.7$ & $0.7$ \\
        \bottomrule
    \end{tabular}
    \caption{Hyperparameters for pre-training models. $P_{\texttt{BLANK}}$ corresponds to the probability of replacing entities with \texttt{[BLANK]}.}
    \label{tab:params}
\end{table}

\begin{table}[t]
\small
    \centering
    \begin{tabular}{l|rrr}
        \toprule
        \textbf{Dataset} &  \textbf{Train} & \textbf{Dev} & \textbf{Test} \\
        \midrule 
        TACRED & 68,124 & 22,631 & 15,509 \\
        SemEval & 6,507 & 1,493 & 2,717 \\
        Wiki80 & 39,200 & 5,600& 11,200   \\
        ChemProt & 4,169 & 2,427& 3,469 \\
        FewRel &44,800 & 11,200& 14,000\\ 
        \bottomrule
    \end{tabular}
    \caption{Numbers of instances in train~/~dev~/~test splits for different RE datasets.}
    \label{tab:data_size}
\end{table}

\paragraph{Pre-training Efficiency}  MTB and our contrastive model have the same architecture as BERT$_{\small \textsc{base}}$~\citep{devlin2019bert}, so they both hold $110$M parameters approximately.
We use four Nvidia 2080Ti GPUs to pre-train models. Pre-training MTB takes $30,000$ training steps and approximately $24$ hours. Pre-training our model takes $3,500$ training steps and approximately $12$ hours. 

\section{RE Fine-tuning} 
\label{appendix:fine-tuning}

\paragraph{RE Datasets} We download TACRED from \texttt{LDC}\footnote{\url{https://catalog.ldc.upenn.edu/LDC2018T24}}, Wiki80, SemEval from \texttt{OpenNRE}\footnote{\url{https://github.com/thunlp/OpenNRE}}, ChemProt from \texttt{sciBert}\footnote{\url{https://github.com/allenai/scibert}}, and FewRel from \texttt{FewRel}\footnote{\url{https://github.com/thunlp/fewrel}}. Table~\ref{tab:data_size} shows detailed statistics for each dataset and Table~\ref{tab:data_size_different} demonstrates the sizes of training data for different supervised RE datasets in 1\%, 10\% and 100\% settings. 
For 1\% and 10\% settings, we randomly sample 1\% and 10\% training data for each relation (so the total training instances for 1\%~/~10\% settings are not exactly 1\%~/~10\% of the total training instances in the original datasets). As shown in the table, the numbers of training instances in SemEval and ChemProt for 1\% setting are extremely small, which explains the abnormal performance.

\begin{table}[t]
\small
    \centering
    \begin{tabular}{l|rrr}
        \toprule
        \textbf{Dataset} &  \textbf{1\%} & \textbf{10}\% & \textbf{100}\% \\
        \midrule 
        TACRED & 703 & 6,833& 68,124\\
        SemEval & 73 & 660& 6,507 \\
        Wiki80 & 400 & 3,920& 3,9200 \\
        ChemProt & 49 & 423& 4,169 \\
        \bottomrule
    \end{tabular}
    \caption{Numbers of training instances in supervised RE datasets under different proportion settings.}
    \label{tab:data_size_different}
\end{table}

\begin{table}[t]
    \small
    \centering
    \begin{tabular}{l|cc}
        \toprule
        \textbf{Parameter} & \textbf{Supervised RE} &  \textbf{Few-Shot RE} \\
        \midrule 
        Learning Rate & $3\times10^{-5}$ & $2\times10^{-5}$\\
        Batch Size & $64$ & $4$ \\
        Epoch & $6$ & $10$ \\
        Sentence Length & $100$ & $128$\\
        Hidden Size & 768 & 768 \\
        \bottomrule
    \end{tabular}
    \caption{Hyperparameters for fine-tuning on relation extraction tasks (BERT, MTB and CP).}
    \label{tab:task_params}
\end{table}

\paragraph{Hyperparameters}

Table~\ref{tab:task_params} shows hyperparameters when finetuning on different RE tasks for BERT, MTB and CP. For CNN, we train the model by SGD with a learning rate of $0.5$, a batch size of $160$ and a hidden size of $230$.
For few-shot RE, we use the recommended hyperparameters in \texttt{FewRel}\footnote{\url{https://github.com/thunlp/FewRel}}.

\paragraph{Multiple Trial Settings}  For all the results on supervised RE, we run each experiment 5 times using 5 different seeds~($42,43,44,45,46$) and select the median of 5 results as the final reported number. For few-shot RE, as the model varies little with different seeds and it is evaluated in a sampling manner, we just run one trial with $10000$ evaluation episodes, which is large enough for the result to converge. We report accuracy (proportion of correct instances in all instances) for Wiki80 and FewRel, and micro \texttt{F$_1$}\footnote{\url{https://en.wikipedia.org/wiki/F1_score}} for all the other datasets.

\end{document}